\documentclass[10pt,letterpaper]{article}
\usepackage[top=0.85in,left=2.75in,footskip=0.75in]{geometry}

\usepackage{amsmath,amssymb}

\usepackage{siunitx}
\usepackage{booktabs}

\usepackage{changepage}

\usepackage{textcomp,marvosym}

\usepackage{float}

\usepackage{cite}

\usepackage{nameref,hyperref}

\usepackage[right]{lineno}

\usepackage[nopatch=eqnum]{microtype}
\DisableLigatures[f]{encoding = *, family = * }

\usepackage[table]{xcolor}

\usepackage{array}

\newcolumntype{+}{!{\vrule width 2pt}}

\newlength\savedwidth

\raggedright
\usepackage[aboveskip=1pt,labelfont=bf,labelsep=period,justification=raggedright,singlelinecheck=off]{caption}

\makeatletter
\renewcommand{\@biblabel}[1]{\quad#1.}
\makeatother

\usepackage{lastpage,fancyhdr,graphicx}
\usepackage{epstopdf}
\fancyheadoffset[L]{2.25in}
\fancyfootoffset[L]{2.25in}
\newenvironment{fullwidthfigure}
  {\begin{figure}[H]\begin{adjustwidth}{-2.25in}{0in}}
  {\end{adjustwidth}\end{figure}}

\begin{document}
\vspace*{0.2in}

\begin{flushleft}
{\Large
\textbf{OpenCap Monocular: 3D Human Kinematics and Musculoskeletal Dynamics from a Single Smartphone Video}
}
\newline
\\
Selim Gilon\textsuperscript{1*},
Emily Y. Miller\textsuperscript{1},
Scott D. Uhlrich\textsuperscript{1,2}
\\
\bigskip
\textbf{1} Department of Mechanical Engineering, University of Utah, Salt Lake City, 84112, United States
\\
\textbf{2} Department of Orthopaedic Surgery, University of Utah, Salt Lake City, 84112, United States
\\
\bigskip

* selim.gilon@utah.edu

\end{flushleft}


%

\section*{Abstract}
Quantifying human movement (kinematics) and musculoskeletal forces (kinetics) at scale—such as estimating quadriceps force during a sit-to-stand movement—could transform the prediction, treatment, and monitoring of mobility-related conditions. However, quantifying kinematics and kinetics traditionally requires costly, time-intensive analysis in a specialized laboratory, limiting clinical translation. Scalable, accurate tools for biomechanical assessment are critically needed. We introduce OpenCap Monocular, an algorithm that estimates 3D skeletal kinematics and kinetics from a single static smartphone video. The method refines 3D human pose estimates from a monocular pose estimation model from computer vision (WHAM) via optimization, computes the kinematics of a biomechanically constrained skeletal model, and estimates kinetics via physics-based simulation and machine learning. We validated OpenCap Monocular against marker-based motion capture and force plate data for walking, squatting, and sit-to-stand tasks. OpenCap Monocular achieved low kinematic error (4.8\textdegree{} mean absolute error [MAE] for rotational degrees of freedom; 3.4 cm MAE for pelvis translations), outperforming a regression-only computer vision baseline by 48\% in rotational accuracy ($p=0.036$) and 69\% in translational accuracy ($p<0.001$). OpenCap Monocular also estimated ground reaction forces during walking with accuracy comparable to, or better than, that of our prior two-camera OpenCap system. We demonstrate that the algorithm estimates important kinetic outcomes with a clinically meaningful level of accuracy in applications related to frailty and knee osteoarthritis, including estimating the knee extension moment during sit-to-stand transitions and the knee adduction moment during walking. OpenCap Monocular is deployed via a smartphone app, a web app, and secure cloud computing (\href{https://opencap.ai}{https://opencap.ai}), enabling free, accessible single-smartphone biomechanical assessments. Such accessibility enables large-scale remote studies and, ultimately, routine evaluations of mobility and function in the clinic or at home. Our code is available at \href{https://github.com/utahmobl/opencap-monocular}{github.com/utahmobl/opencap-monocular}.


%
\section*{Author Summary}
The ability to easily measure human movement and musculoskeletal forces has the potential to improve the treatment of movement-related disorders. However, precise biomechanical analysis has traditionally required costly, time-consuming laboratory analyses, limiting its impact on clinical practice. To address this gap, we developed OpenCap Monocular, an open-source tool that estimates 3D skeletal motion and musculoskeletal forces using video recorded with a single smartphone.

Our algorithm refines motion estimates from computer vision models—which often contain physically implausible artifacts such as foot sliding—by integrating physics-based modeling with machine learning to generate physically and biomechanically consistent motion and force estimates. The system estimates 3D human motion more accurately than a single-camera computer vision model alone and achieves accuracy comparable to more cumbersome two-camera setups.

Its force estimates achieve clinically meaningful accuracy, capturing knee loading metrics related to osteoarthritis and muscle force patterns linked to age-related declines in physical function. By packaging this complex pipeline into a free, automated cloud application, OpenCap Monocular enables clinicians and researchers to conduct precise, large-scale movement studies in clinics and homes using equipment they already carry in their pockets.


\section{Introduction}

Quantitative analysis of human movement provides critical information across fields such as rehabilitation, sports science, ergonomics, and the treatment of musculoskeletal and neuromuscular disorders. Measures of movement kinematics (e.g., joint angles, velocities) and kinetics (e.g., joint moments, ground reaction forces, muscle forces) can predict the risk of injury and disease progression, track functional recovery, and evaluate the efficacy of interventions \cite{Paterno, Clark_2022, Diagnosing_Neurodegenerative, Isabel_2023, Alderink_2025, Collins_2018, Uhlrich_lancet}.

Traditionally, the gold standard for high-fidelity movement analysis has been laboratory-based motion capture \cite{stebbins}. This approach uses specialized cameras to track reflective markers on the body and force plates to measure ground reaction forces. While accurate, this method requires expensive equipment (often $>$\$150,000), dedicated laboratory space, specialized expertise for data collection and processing, and considerable time investment (often hours to days per participant) \cite{opencap}. Thus, it is rarely used clinically. Although widely used in research, the limited scalability of motion capture has constrained our ability to study movement in large cohorts in clinical, community, or home settings (Fig~\ref{fig:moti}). Most biomechanics studies are limited to laboratory settings and include a median of only 12 to 21 participants \cite{Oliveira, Duane}.

Mobile sensing techniques for measuring human movement can address these challenges, but scalability and accuracy challenges remain. Inertial measurement units can estimate kinematics outside the lab. However, they require the donning and doffing of up to 15 sensors to estimate whole-body kinematics and kinetics \cite{Weygers, OpenSense, Konrath, Karatsidis}. Wearable sensors are well-suited for long-term monitoring of specific biomechanical variables \cite{Servais, Wang, OpenSense}. However, they are impractical for rapid, routine assessments of whole-body movement in large-scale, decentralized studies or in clinical practice (Fig~\ref{fig:moti}).

\begin{fullwidthfigure}
\includegraphics[width=\linewidth]{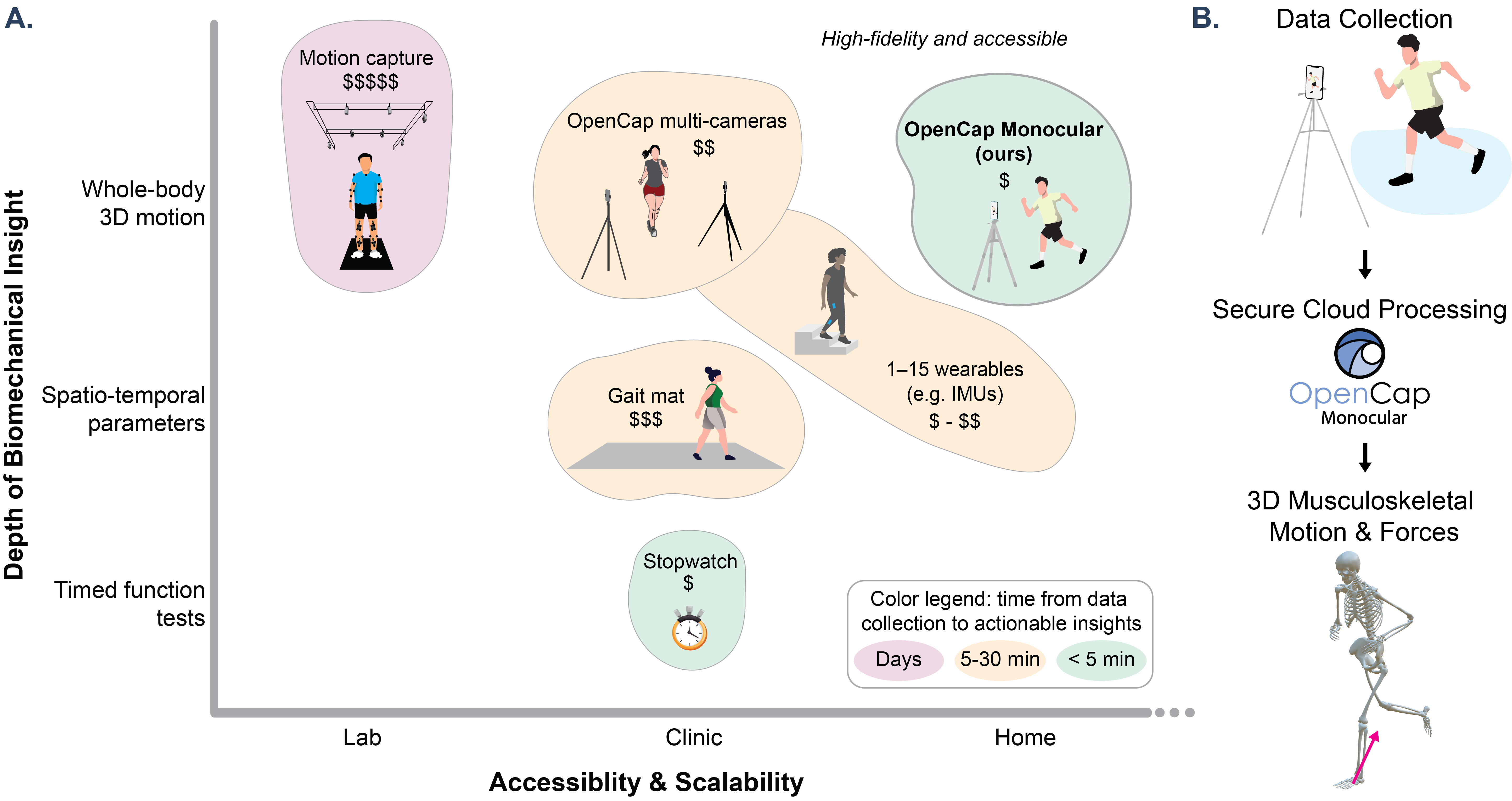}
\caption{\textbf{OpenCap Monocular Enables Scalable Evaluation of 3D Human Motion and Musculoskeletal Dynamics}. 
(\textbf{A.}) Traditional lab-based motion capture provides valuable, high-fidelity biomechanical assessments, but it is costly and time-consuming. Clinical assessments of function, such as timed functional tests, fail to capture the nuances of full-body biomechanics. OpenCap Monocular addresses the need for fast, scalable, and accurate tools to quantify whole-body motion. This software enables 3D biomechanical assessments in large-scale, ecologically valid studies and supports integration into routine clinical practice.
(\textbf{B.}) OpenCap Monocular enables 3D assessment of kinematics and kinetics with a single smartphone. The pipeline is freely available through our mobile and web applications and secure cloud processing infrastructure.}
\label{fig:moti}
\end{fullwidthfigure}

Video-based approaches, leveraging advancements in computer vision and deep learning for human pose estimation \cite{desmarais, openpose, hrnet, kanko}, are a promising avenue for large-scale, rapid measurement of whole-body motion, due to the ubiquity of smartphone cameras \cite{melissa}. We previously developed OpenCap, an open-source platform that quantifies 3D kinematics and musculoskeletal dynamics from two or more smartphone videos \cite{opencap}. Multi-camera OpenCap enables motion to be measured in approximately 10 minutes with equipment that costs $<$\$1,000. We deployed the open-source software using cloud computing and a freely available web application. Deploying these algorithms into an easy-to-use application has enabled 14,000 researchers to collect 400,000 motion trials in the 3 years since its release. Multi-camera video-based systems, like OpenCap, have enabled large-scale studies of movement in more ecologically valid settings (e.g., in campus gymnasiums or at patient advocacy conferences) that would have been infeasible with the lab-based approach \cite{Ruth, Gurchiek}. However, even the requirement of multiple calibrated, tripod-mounted smartphones plus a laptop is a barrier in specific contexts, particularly for regular clinical or in-home assessments. The ability to perform accurate 3D motion capture and dynamic analysis from a \textit{single} smartphone video would represent a significant improvement in accessibility, potentially empowering billions of smartphone users worldwide with tools for quantitative movement assessment.

Here, we developed OpenCap Monocular, an open-source, cloud-deployed algorithm for estimating 3D skeletal kinematics and kinetics from a single static smartphone video (Fig~\ref{fig:moti}). We present validation against gold-standard marker-based motion capture and force plate measurements. We hypothesized that OpenCap Monocular would result in lower kinematic and kinetic errors compared to directly applying inverse kinematics to computer vision model outputs (i.e., CV + IK). We also compare accuracy to the original two-camera OpenCap platform \cite{opencap}. We then evaluate OpenCap Monocular's utility for two downstream clinical tasks. First, we analyze the kinetics of a sit-to-stand transition, an activity that reflects age-related reductions in quadriceps strength \cite{heijden, Moreland, van_der_kruk}. We hypothesize that OpenCap Monocular can detect a redistribution of lower-extremity joint moments from the knee to the ankle and the hip during chair rise with a quadriceps-avoidance strategy. We further test whether errors in OpenCap Monocular's knee extension moment estimates fall below a clinically meaningful threshold of 11 Nm---the difference observed between individuals with and without early signs of frailty (pre-frailty) \cite{frailty_seko2024}. In a second clinical use case, we evaluate OpenCap Monocular's accuracy in estimating the knee adduction moment during walking, a key dynamic metric of knee loading associated with medial compartment knee osteoarthritis progression \cite{Miyazaki, Amin}. We hypothesize that OpenCap Monocular can estimate the knee adduction moment with errors below a clinically meaningful threshold of 0.5\% bodyweight (BW)·height (ht) \cite{Miyazaki, Amin}, supporting its use for predicting progression and evaluating joint-offloading interventions.

\section{Methods}

\subsection{Data Collection with OpenCap Monocular}
\label{sec:data-collection}
Setting up an OpenCap Monocular recording takes less than one minute and requires minimal equipment: one iPhone or iPad and a tripod. No markers, calibration frames, or force plates are needed. After logging into our HIPAA-compliant web application, users can either (1) control data collection directly on the tripod-mounted iOS device that is recording video or (2) use a separate internet-connected device (e.g., a laptop) to control the recording iOS device. The captured video is automatically uploaded to the cloud for processing, where kinematics are computed in under two minutes for a 10-second video. Results can be visualized directly in the web application or downloaded through the application programming interface for further analysis (e.g., kinetics). The iOS app and web platform are freely available at \href{https://opencap.ai/}{https://opencap.ai/}.

\subsection{OpenCap Monocular Pipeline: Static Video to Kinetics}
\label{sec:pipeline-steps}
Fig~\ref{fig:process} illustrates the five core processing steps: initial 3D pose estimation, pose refinement optimization, marker extraction from SMPL vertices, inverse kinematics, and estimation of musculoskeletal dynamics. The first four steps are automated in the cloud. For walking trials, the fifth step, estimation of ground reaction forces, is available via a machine learning model \cite{tan2025gaitdynamics, Miller2025}. For all trial types, we provide code that performs this step and the musculoskeletal dynamics pipeline offline in post-processing \cite{opencap, falisse_rapid}.

\begin{fullwidthfigure}
\includegraphics[width=\linewidth]{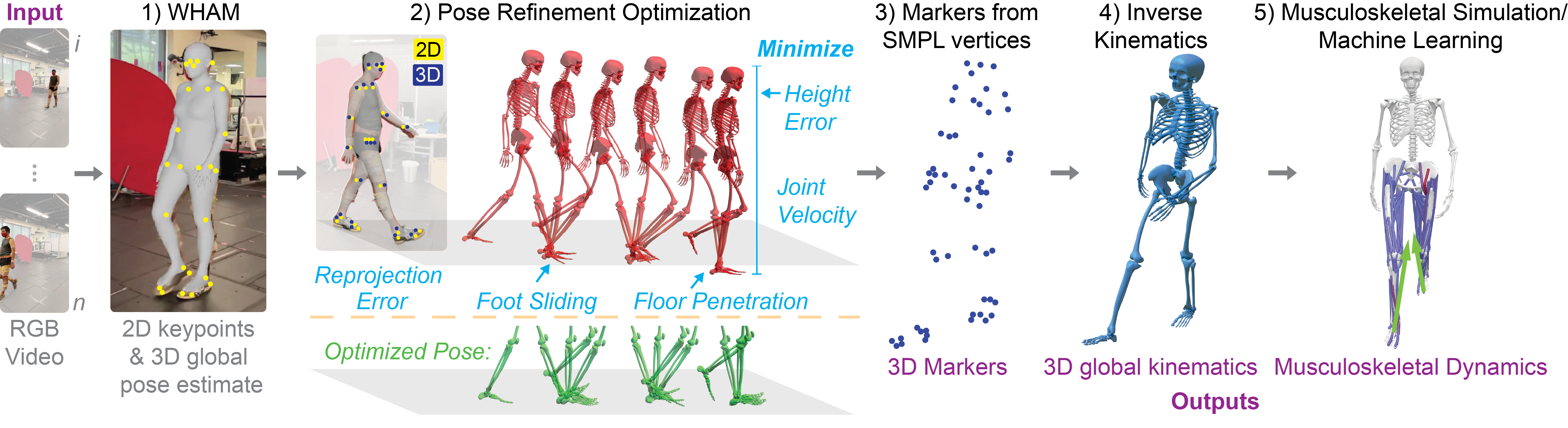}
\caption{\textbf{OpenCap Monocular Algorithm}. OpenCap Monocular estimates 3D global kinematics and kinetics from a single, static smartphone video. (1) Computer vision models, ViTPose \cite{vitpose} and WHAM \cite{WHAM}, estimate 2D keypoints and an initial 3D human global pose, represented by a sequence of SMPL model parameters~\cite{loper}. (2) This initial pose sequence (top, red skeleton) often contains physical inaccuracies like translational drift and foot-floor penetration. To correct this, we apply a pose-refinement optimization that minimizes reprojection error, foot sliding/penetration, and excessive joint velocity. The output is a more physically plausible, optimized pose sequence (bottom, green skeleton). (3) A set of virtual skin markers is extracted from the vertices of the refined SMPL mesh and (4) tracked with OpenSim Inverse Kinematics~\cite{opensim} to obtain 3D joint kinematics. (5) Physics-based and machine learning algorithms are used to estimate kinetics (e.g., ground reaction and muscle forces) from the monocular kinematics, without the need for force plates~\cite{tan2025gaitdynamics, opencap, Miller2025}.} 
\label{fig:process}
\end{fullwidthfigure}

\subsubsection{Initial 3D Pose Estimation}
We first use WHAM \cite{WHAM} to estimate the global 3D human pose, which is a sequence of SMPL \cite{loper} model parameters, including body shape ($\beta_{0}$), body pose ($\theta_{0}$), global translation ($\tau_{0}$), and global orientation ($\Gamma_{0}$). WHAM also provides an estimate of the camera's extrinsic parameters, $\xi$. Additionally, we use ViTPose \cite{vitpose} to estimate 2D keypoint locations and confidence scores for both the WHAM pipeline and subsequent optimization steps. WHAM also provides ground contact probabilities for the heel and toe, which are used to guide our subsequent refinement step. While WHAM provides a strong initial estimate of the 3D motion, it can suffer from inaccuracies such as translational drift and physically implausible foot-floor interactions (e.g., sliding, penetration), which motivate our pose refinement step.

\subsubsection{Pose Refinement Optimization}
\label{sec:poserefinement}
To improve the physical plausibility of the initial pose estimates, we implemented a two-stage optimization procedure that refines camera extrinsic parameters and SMPL pose and shape parameters, ensuring consistency with observed 2D keypoints and the physical constraints of human movement. The optimization problem is formulated in PyTorch, enabling automatic differentiation and GPU acceleration, and both stages optimize all design variables simultaneously over the whole sequence. We assume that the camera is not moving, that the body shape is constant over time, and that the individual's height is known (it is queried during recording in the web application). For videos captured with our iOS application, camera intrinsics are known a priori via a database of intrinsic parameters for all iOS devices released since 2018.

In the first optimization stage, we solve for ($\beta$) and the camera extrinsic ($\xi$) parameters while holding the global pose ($\theta_{0}$, $\tau_{0}$, $\Gamma_{0}$) constant. The objective function is described in Eq~(\ref{eq:stage1}):
\begin{equation}\label{eq:stage1}
\mathcal{J}_{\mathrm{stage1}} = w_{\mathrm{r}}\,L_{\mathrm{repr}} + w_{h}\,L_{\mathrm{height}} + w_{\beta}\,L_{\beta}
\end{equation}
where:

$L_{\mathrm{repr}}$ is the confidence-weighted 2-D reprojection error:
\begin{equation}
L_{\mathrm{repr}}(\xi, K) = \sum_{i=1}^{N} w_i \, \big\lVert \Pi(\mathbf{X}_i^{3D}; K,\xi) - \mathbf{x}_i^{2D} \big\rVert_2^2
\end{equation}

\noindent where:  

$N$ is the number of keypoints,  
$w_i$ is the confidence weight of keypoint $i$,  
$\mathbf{X}_i^{3D}$ is the 3D position of keypoint $i$,  
$\Pi(\mathbf{X}_i^{3D}; K, \xi)$ denotes the projection of the 3D point into image coordinates using the camera intrinsics matrix $K$ and extrinsic parameters $\xi$ (rotation and translation), and  
$\mathbf{x}_i^{2D}$ is the observed 2D keypoint in the image.  

$L_{\mathrm{height}}=\bigl(\hat{h}-h_{\mathrm{}})^{2}$ 
penalizes
deviations from the individual's known height $\hat{h}$. 

And $L_{\beta}^{\mathrm{reg}}$ penalizes deviations in body-shape parameters from the initial WHAM estimate $\beta_{0}$. 
\begin{equation}
L_{\beta}^{\mathrm{reg}} = \sum_{j=1}^{D} \bigl(\beta_j - \beta_{0,j}\bigr)^{2}
\end{equation}

\noindent where: 

$D$ is the dimensionality of the body‐shape parameter vector (10),
$\beta_j$ is the $j$-th body‐shape parameter, and
$\beta_{0,j}$ is the corresponding WHAM estimate for parameter $j$. 

In the second optimization stage, the global pose ($\theta$, $\tau$, $\Gamma$) and camera extrinsics are refined while the body shape ($\beta$) is held constant. The objective function is described in Eq~(\ref{eq:stage2}).
\begin{align}\label{eq:stage2}
\mathcal{J}_{\mathrm{stage2}} &= w_{\mathrm{r}}\,L_{\mathrm{repr}}
             + w_{\mathrm{c}}\,L_{\mathrm{cam}}+ w_{v}\,L_{\mathrm{foot\,vel}}\\
             &+ w_{s}\,L_{\mathrm{foot\,slide}}
             + w_{f}\,L_{\mathrm{flat}} 
             + w_{sm}\,L_{\mathrm{smooth}} \notag
\end{align}

where: 
\begin{itemize}
  \item $L_{\mathrm{cam}}$ penalizes deviations in camera extrinsics from stage 1,
  \item $L_{\mathrm{foot\,vel}}$ penalizes non-zero velocities of heel and toe markers during contact,
  \item $L_{\mathrm{foot\,slide}}$ penalizes movement (variance in position) of the heel and toe during bouts of continuous contact,
  \item $L_{\mathrm{flat}}$ enforces a consistent vertical position of heel and toe markers during contact events across the sequence,
  \item $L_{\mathrm{smooth}}$ penalizes joint linear velocity, encouraging smoothness.
\end{itemize}

Mathematical expressions for each term in the second-stage objective function are provided in our open-source implementation for full reproducibility.

We tuned the weights of the second optimization stage using previously published OpenCap datasets \cite{opencap}, based on 3D marker errors relative to marker-based motion capture and qualitative inspection of movement plausibility \cite{opencap}. At the time of publication, the deployed pipeline uses separate, activity-specific weight sets for gait, squatting, sit-to-stand, and other movements. To automatically select the appropriate parameter set at runtime, we employ a video-understanding foundation model (Video-LLaMA3 \cite{llama}) to classify the activity being performed. This design improves robustness across diverse recording environments and enables extensible activity classification without manual intervention or additional optimization. The parameters used are listed in Supplementary Table S1.

\subsubsection{Marker Extraction}
\label{sec:markers_extraction}
For downstream kinematic and kinetic processing using a biomechanical model in OpenSim, we extract 38 virtual surface marker positions as vertices of the SMPL mesh. These markers characterize the motion of the forearm, upper arm, torso, pelvis, thigh, shank, and foot segments. We extract a 'static' set of markers using a default standing SMPL pose to scale a biomechanical model. Then, we extract marker trajectories during the motion sequence using the optimized global pose ($\theta_{opt2}$, $\tau_{opt2}$, $\Gamma_{opt2}$).

\subsubsection{Inverse Kinematics of a Musculoskeletal Model}
\label{sec:inverse_kinematics}
We use a musculoskeletal model \cite{opencap, laiarnold, rajagopal, opensim} with 33 degrees of freedom (6 for the pelvis in the ground, 3 for the lumbar, 3 for each hip, 1 for each knee, 2 for each ankle, 3 for each shoulder, 2 for each elbow, and 1 metatarsophalangeal joint per foot [unlocked for physics simulation only]). Importantly, unlike the SMPL model, the joints in this constrained OpenSim model can only move in biomechanically plausible ways. For example, whereas the SMPL model represents the knee with three rotational degrees of freedom, our biomechanical model includes six axes of motion, with five constrained as spline functions of the single flexion–extension degree of freedom.

For each motion trial, we scale the musculoskeletal model to the individual's anthropometry using the OpenSim Scale tool \cite{opensim} with markers extracted from the static-posed SMPL model. We then solve for the kinematics of this scaled model using the marker trajectories during the motion and the Inverse Kinematics (IK) tool. The resulting monocular kinematics are trajectories of biomechanically plausible pelvis translations and joint kinematics.

\subsubsection{Musculoskeletal Dynamics from Monocular Kinematics}
Here we employ two different approaches to estimate musculoskeletal dynamics (i.e., kinetics) from the monocular kinematics: physics-based simulation \cite{opencap,falisse_rapid} and machine learning \cite{tan2025gaitdynamics}.

The physics-based approach uses the simulation methods described in Uhlrich, Falisse, and Kidzinski et al. (2023) \cite{opencap}. Briefly, we estimate musculoskeletal dynamics (ground reaction forces, joint moments, and muscle forces) using a torque or muscle-driven dynamic simulation that tracks the monocular kinematics, without the need for experimental force plate data. We model foot-floor contact using six smooth Hunt-Crossley contact spheres per foot \cite{Serrancoli, Simbody}. The simulation is posed as an optimal control problem, formulated using direct collocation with CasADI \cite{CasADi}, and solved using IPOPT with algorithmic differentiation (OpenSimAD) for gradient computation \cite{falisse_rapid, wachter}. The optimization solves for kinematics, muscle excitations, and torque actuator controls that track the monocular kinematics while minimizing the sum-squared muscle activations, subject to constraints on muscle and skeletal dynamics. The resulting kinematics and kinetics closely track the input motion, but are dynamically consistent (i.e., no residual forces or moments at the ground-pelvis joint), and could plausibly be generated by muscles. Here, we use muscle-driven simulations to obtain dynamics for the sit-to-stand activity.

Since the release of the original multi-camera OpenCap platform \cite{opencap}, a machine learning model that predicts ground reaction forces during gait from kinematics has been developed (GaitDynamics) \cite{tan2025gaitdynamics}. This model improves the speed and accuracy of ground force prediction from kinematics during walking compared to physics-based simulations alone. To estimate walking kinetics in this study, we use GaitDynamics to predict ground forces, and we track these predictions in a physics simulation to estimate joint moments \cite{Miller2025}.

\subsection{Secure Cloud Deployment}
The OpenCap Monocular pipeline is integrated into the OpenCap web and iOS applications (Fig. S2). Videos uploaded to the cloud are queued and processed by GPU servers. A 10-second video takes less than two minutes to process using an NVIDIA RTX 4090 GPU. Once processed, the results can be visualized, further analyzed (e.g., automated gait analysis), and downloaded from the web application. Alternatively, data can be downloaded programmatically for further processing (e.g., estimating kinetics) using our application programming interface and post-processing software. The codebase is open source, and cloud processing is provided free of charge to the research community.

\subsection{Validation Protocol}
\label{sec:validation}
To validate our method, we used the publicly available dataset of synchronized video, marker-based motion capture, and force plate recordings from the multi-camera OpenCap study \cite{opencap}; full experimental details are provided in the prior publication. Briefly, ten healthy adults (5 females; age 26±4 years; mass 74±8 kg) performed several activities, including level walking, five bodyweight squats, and five sit-to-stand transitions. To simulate movement patterns relevant to clinical populations, participants were also instructed to perform modified versions of these tasks, such as squats while offloading one foot and sit-to-stand transitions with increased trunk flexion and angular velocity to simulate a strategy commonly observed in older adults with quadriceps weakness \cite{Moreland, heijden}. The dataset also includes variations of gait, in which participants were instructed to walk with a trunk-sway modification to emulate compensatory movement patterns.
Thus, our kinematic comparison for each individual comprises 10 squats, 10 sit-to-stand transitions, and six walking trials, with varied kinematic patterns for each activity. Participants whose faces are visible in images and videos in this article (e.g., Supplementary Video and Fig. S2) have provided informed consent to the sharing of identifiable video data, through a protocol approved by the University of Utah Institutional Review Board.

For the OpenCap Monocular analysis, we used the 45\textdegree{} anterolateral camera view from the original multi-camera OpenCap recordings. The frontal and sagittal-only camera views produced visually worse 3D monocular kinematics due to prolonged segment occlusion and a lack of information in an entire plane of movement. We compared OpenCap Monocular kinematics with marker-based motion capture, a computer vision baseline (CV + IK), and the previously published two-camera OpenCap algorithm. The CV + IK baseline used the SMPL model prediction directly from WHAM, bypassed the pose refinement optimization step (Section~\ref{sec:poserefinement}), and proceeded to the marker extraction and IK steps (Section~\ref{sec:markers_extraction}, Section~\ref{sec:inverse_kinematics}). We compute kinematic accuracy as a mean absolute error (MAE) across three translational degrees of freedom (pelvis translations) and 18 rotational degrees of freedom: ankles (4), knees (2), hips (6), pelvis global orientation (3), and lumbar (3).
We compared all-activity kinematic errors between the computer vision baseline and OpenCap Monocular using paired t-tests ($\alpha = 0.05$, $n=10$). Ground reaction force accuracy was quantified as the mean absolute error (MAE) during the stance phase for each directional component, normalized by body weight. Walking ground reaction force errors were compared between methods using paired t-tests ($\alpha = 0.05$, $n=10$). All statistical analyses were conducted in Python (v3.9.21) using SciPy (v1.11.4).

\subsection{Clinical Use Case 1: Joint Moments during Chair Rise}
\label{sec:validation_quad}
To demonstrate the clinical utility of OpenCap Monocular, we analyzed the sit-to-stand (STS) movement, a fundamental functional assessment in clinical practice. Rising strategies vary with age and are associated with distinct muscle force requirements \cite{smith}. Older adults often increase trunk flexion when rising from a chair, shifting muscular demand from the knee extensors to the hip extensors and ankle plantarflexors \cite{van_der_kruk}. This compensatory strategy is associated with reduced functional strength \cite{heijden} and an increased risk of falls \cite{Moreland}.
We used the OpenCap validation dataset \cite{opencap}, in which 10 healthy individuals performed sit-to-stand transitions under two conditions: 5 repetitions using their natural strategy, followed by 5 repetitions with deliberately increased trunk flexion. Joint moment values were averaged over the rising phases of the three central repetitions.
We first evaluated the accuracy of the knee extension moment, averaged over the rising phase, a proxy for quadriceps force. We compared this accuracy to an 11 Nm clinically relevant accuracy threshold, as this is the average difference in moment between adults with and without early frailty \cite{frailty_seko2024}. We also performed one-sample t-tests on the changes in knee, hip, and ankle moments between the natural and trunk-flexion conditions to test whether OpenCap Monocular could detect group changes in joint moments similarly to motion capture and force plates.

\subsection{Clinical Use Case 2: Knee Loading during Walking}
\label{sec:validation_kam}
To further demonstrate the clinical relevance of OpenCap Monocular, we evaluated the accuracy of the knee adduction moment (KAM), a loading metric associated with the onset and progression of medial compartment knee osteoarthritis \cite{Miyazaki, Amin}. Elevated KAM values during walking indicate increased medial compartment loading, which accelerates cartilage degeneration \cite{Miyazaki, Bennell}. Accurate, scalable estimation of the KAM from smartphone videos could therefore enable remote assessment of knee joint loading in large populations and facilitate early detection, monitoring, and personalized rehabilitation for medial knee osteoarthritis \cite{Uhlrich_lancet}.
We used walking trials from the OpenCap validation dataset \cite{opencap}. The KAM was computed from monocular kinematics using a hybrid pipeline that combines physics-based and machine-learning approaches \cite{tan2025gaitdynamics, Miller2025}. We compared these estimates to gold-standard inverse dynamics derived from motion capture and force plates. We focused on the first peak of the KAM during the stance phase, due to its relationship to disease progression  \cite{Miyazaki}. 
We computed the mean absolute error (MAE) of the first-peak KAM, normalized to body weight and height, and compared it against a clinically meaningful threshold of 0.5\% BW·ht. This threshold has been shown to differentiate individuals with slow versus rapid medial knee osteoarthritis progression \cite{Miyazaki, Mundermann_1, Mundermann_2, Amin, Boswell} and represents the lower bound of the clinically relevant range (0.5–2.2\% BW·ht) used for OA diagnosis and progression risk assessment.

\section{Results}

\subsection{Kinematic Accuracy}
\label{sec:kinematic-acc}
Across all activities, OpenCap Monocular had an MAE of $4.8^{\circ}$ for rotational kinematics and 3.4 cm for translational kinematics, compared to marker-based motion capture. These errors are $4.5^{\circ}$ (48\%, $p=0.036$) and 7.6 cm (69\%, $p<0.001$) lower than the computer vision baseline (CV+IK), highlighting the importance of the pose refinement step. OpenCap Monocular's rotational accuracy was within 1\textdegree{} of two-camera OpenCap, and translational accuracy within 2 cm (Fig~\ref{fig:kinematics}). Rotational accuracy by degree of freedom is provided in Fig. S1.

\begin{fullwidthfigure}
\includegraphics[width=\linewidth]{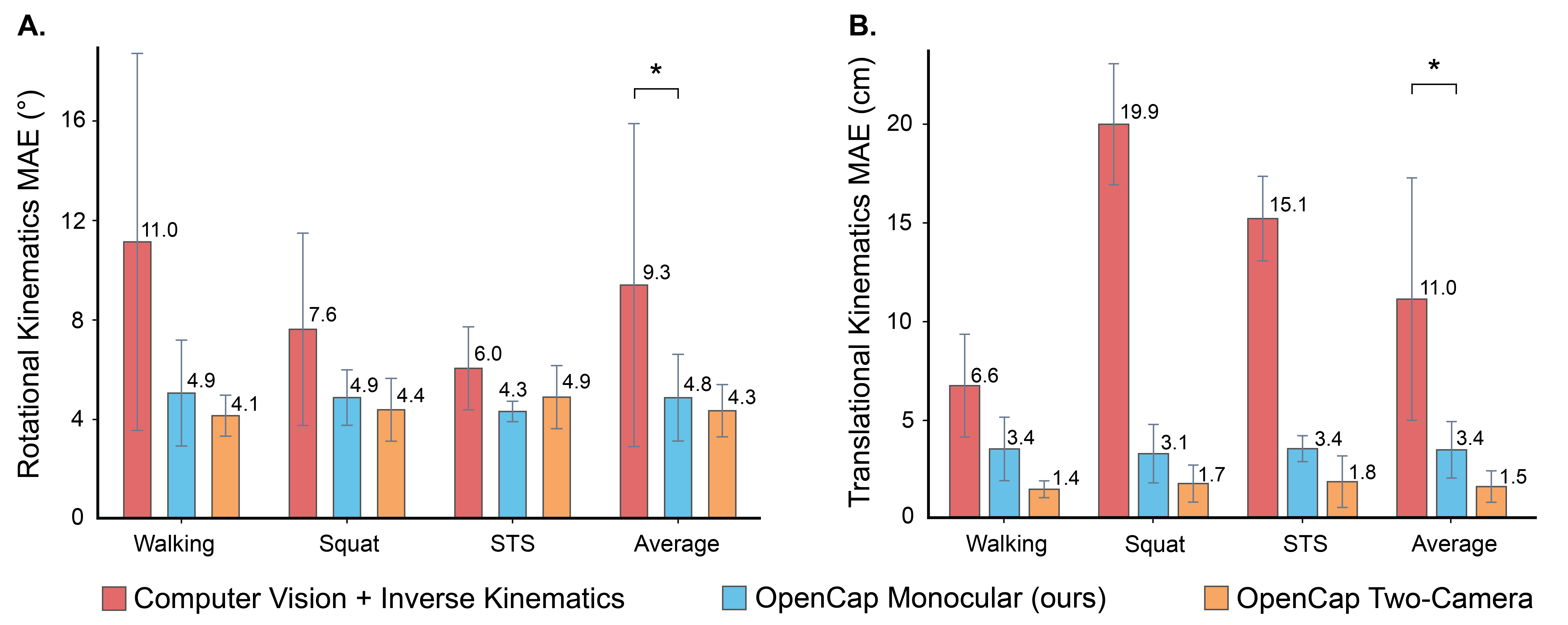}
\caption{\textbf{Kinematic Accuracy}. The mean (bar) and standard deviation (error bar) of mean absolute errors (MAE) in kinematics across activities (STS stands for sit-to-stand), compared to marker-based motion capture. * indicates $p < 0.05$.
Compared to the computer vision baseline model, OpenCap Monocular demonstrated (\textbf{A}) 48\% lower errors across 18 rotational degrees of freedom ($p = 0.036$) and \textbf{(B)} 69\% lower errors across three pelvic translational degrees of freedom ($p < 0.001$), averaged across activities.
}
\label{fig:kinematics}
\end{fullwidthfigure}

In addition to improving accuracy (Fig~\ref{fig:kinematics}), the pose refinement optimization step in the OpenCap Monocular reduced translational drift (Fig~\ref{fig:drift_sts}). Translations from CV+IK often drifted over time. For example, after five repetitions of the sit-to-stand, the CV+IK pelvis drifted by an average of 56.9 cm, whereas our refined pose remained more stable with an average pelvis translational error after five repetitions of 4.9 cm (Fig~\ref{fig:drift_sts}).

\begin{fullwidthfigure}
\includegraphics[width=\linewidth]{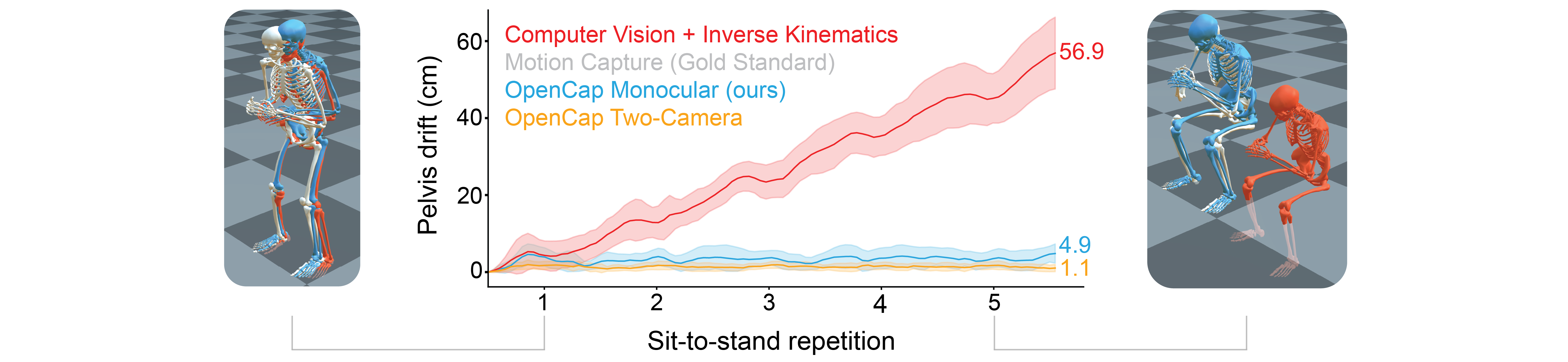}
\caption{\textbf{Impact of Pose Refinement on Translational Drift}. The mean (line) and standard deviation (shading) of pelvis translational drift (Euclidean distance between the estimated pelvis position and marker-based motion capture) over five sit-to-stand repetitions. All pelvis origins were aligned at the initial time point. OpenCap Monocular drifted an order of magnitude less than the computer vision plus inverse kinematics baseline (CV+IK) but still more than the two-camera OpenCap approach, which can compute depth analytically. Representative skeletal kinematics are shown during the first and fifth repetitions for marker-based motion capture (white), OpenCap Monocular (blue), and CV+IK (red).}
\label{fig:drift_sts}
\end{fullwidthfigure}

\subsection{Kinetic Accuracy}
OpenCap Monocular with the GaitDynamics model estimated ground reaction forces with an MAE of 9.7\% BW compared to force plate measurements. This outperforms the CV + IK baseline of 13.6\% BW (58\% improvement) in the vertical direction ($p=0.002$; Fig~\ref{fig:grf_walking}). Although not statistically tested, OpenCap Monocular yields slightly lower vertical ground reaction force errors than estimates obtained from two-camera OpenCap kinematics, using either physics-based simulation or the GaitDynamics model (12.2--13.5\% BW). 

\begin{fullwidthfigure}
\includegraphics[width=\linewidth]{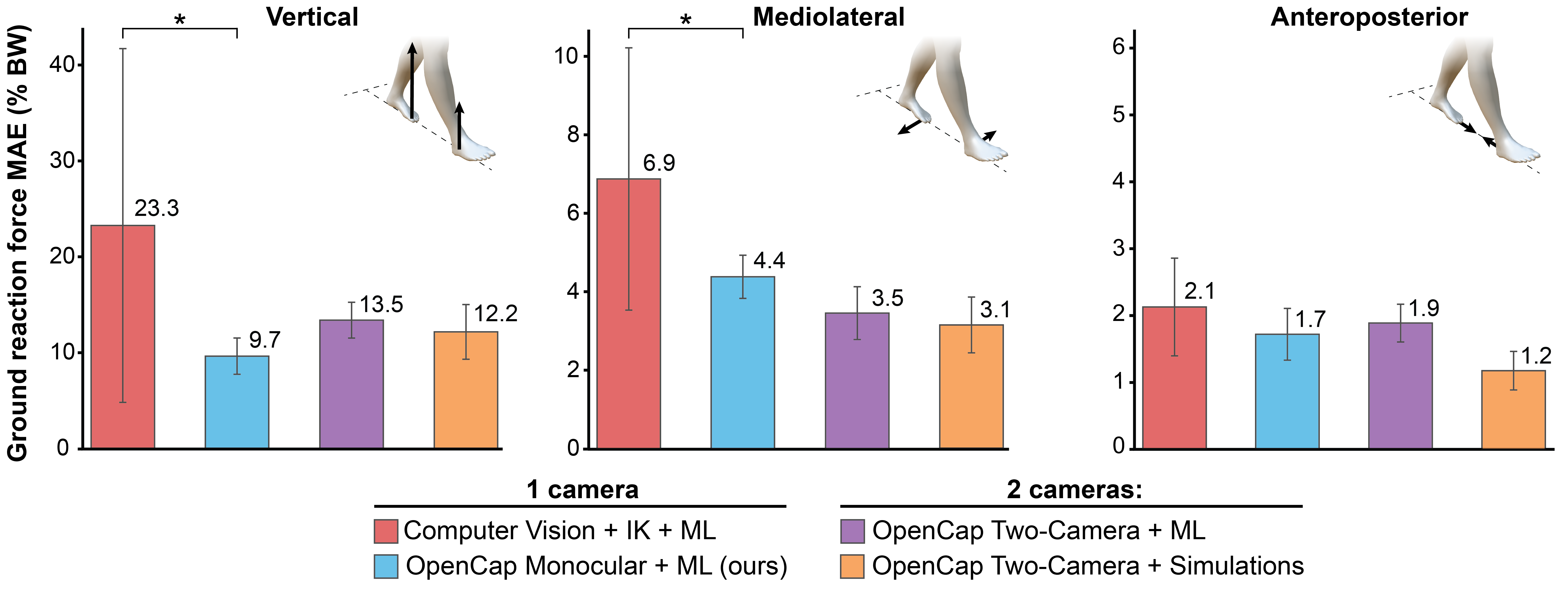}
\caption{\textbf{Ground Reaction Force Accuracy}. The mean (bar) and standard deviation (error bar) of mean absolute errors (MAE) in ground reaction forces during walking compared to force plates. OpenCap Monocular kinematics coupled with the GaitDynamics \cite{tan2025gaitdynamics} machine learning (ML) model estimated ground reaction forces more accurately (vertical: $p=0.002$; mediolateral: $p=0.002$; anteroposterior: $p=0.065$) than the baseline computer vision model (CV+IK) and GaitDynamics (* indicates $p < 0.05$). We also compare to forces derived from two-camera OpenCap kinematics with either physics-based simulation \cite{opencap} or GaitDynamics (ML). Interestingly, OpenCap Monocular + ML yielded slightly lower vertical force errors than either two-camera approach, despite using only one camera, potentially due to improved vertical center-of-mass kinematics from OpenCap Monocular's pose refinement step.}
\label{fig:grf_walking}
\end{fullwidthfigure}

\subsection{Clinical Use Case 1: Joint Moments during Chair Rise}
\label{sec:results_sts}
OpenCap Monocular detected the redistribution of lower-extremity joint moments from a normal to an exaggerated trunk-lean sit-to-stand movement. It detected a reduction in knee extension moment ($p=0.015$), 
an increase in hip extension moment ($p=0.003$), 
and an increase in ankle plantarflexion moment ($p=0.044$). 
Importantly, the direction of these changes matched inverse dynamics analysis using gold-standard motion capture and force-plate systems 
($p=0.027$; 
$p=0.013$; 
$p=0.004$, respectively).

Across all sit-to-stand trials, the rising phase--averaged knee extension moment estimated by OpenCap Monocular showed strong agreement with motion capture ($r^{2}=0.64$), with an MAE of 5.8~Nm, which is below the 11~Nm clinically meaningful threshold related to pre-frailty \cite{frailty_seko2024}.

\begin{fullwidthfigure}
\includegraphics[width=\linewidth]{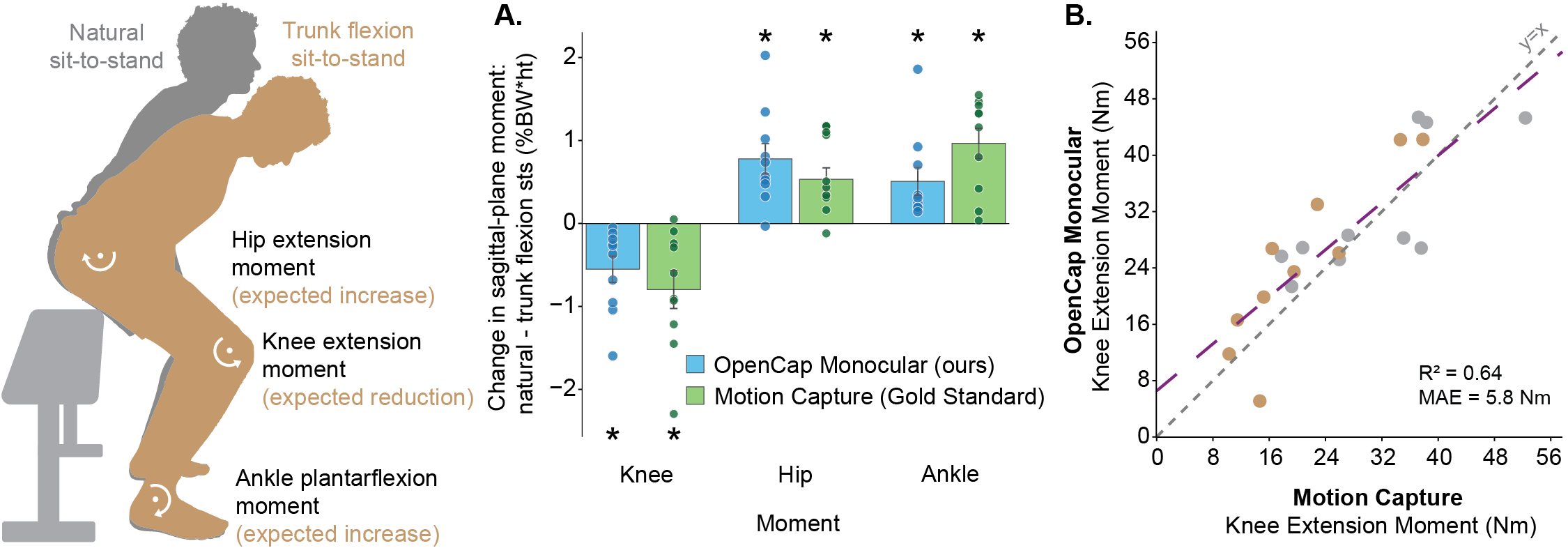}
\caption{\textbf{Clinical Use Case 1: Detecting Joint Moment Differences during a Quadriceps-Avoidance Sit-to-Stand Transition}. Ten participants completed the Five Times Sit-to-Stand test naturally and with instruction to increase their trunk flexion angle and angular velocity during lift-off, a compensatory strategy often used by individuals with quadriceps weakness to shift demand from the knee extensors (quadriceps) to the hip extensors and ankle plantarflexors. (\textbf{A.}) Changes (mean ± standard deviation) in lower-extremity joint moments, averaged over the standing phase, from the natural to the trunk flexion condition, normalized to bodyweight (BW) and height (ht). OpenCap Monocular detected the expected reduction in knee extension moment and increase in hip and ankle moments ($p=0.015-0.044$), similar to motion capture and force plates ($p=0.004-0.027$). * indicates $p < 0.05$.
(\textbf{B.}) OpenCap Monocular estimated the rising phase--averaged knee extension moment with 5.8 Nm of mean absolute error (MAE), compared to motion capture and force plates. This falls below an 11 Nm clinically meaningful threshold that differentiates older adults with and without early signs of frailty \cite{frailty_seko2024}.}
\label{fig:sts}
\end{fullwidthfigure}

\subsection{Clinical Use Case 2: Knee Loading during Walking}
\label{sec:results_kam}
OpenCap Monocular accurately estimated the knee adduction moment (KAM) during walking, demonstrating close agreement with motion capture and force plate--derived inverse dynamics (Fig~\ref{fig:KAM}). The estimated KAM waveform captured both characteristic KAM peaks during the stance phase, with timing and magnitudes similar to those of the gold standard. The MAE of the first peak KAM from OpenCap Monocular's hybrid kinetics pipeline (\(0.36\%~\mathrm{BW{\cdot}ht}\)) was comparable to the two-camera OpenCap approach (\(0.41\%~\mathrm{BW{\cdot}ht}\)) and was below the clinically meaningful threshold of \(0.5\%~\mathrm{BW{\cdot}ht}\)~\cite{Miyazaki, Amin, Mundermann_1, Mundermann_2}. 

\begin{fullwidthfigure}
\includegraphics[width=\linewidth]{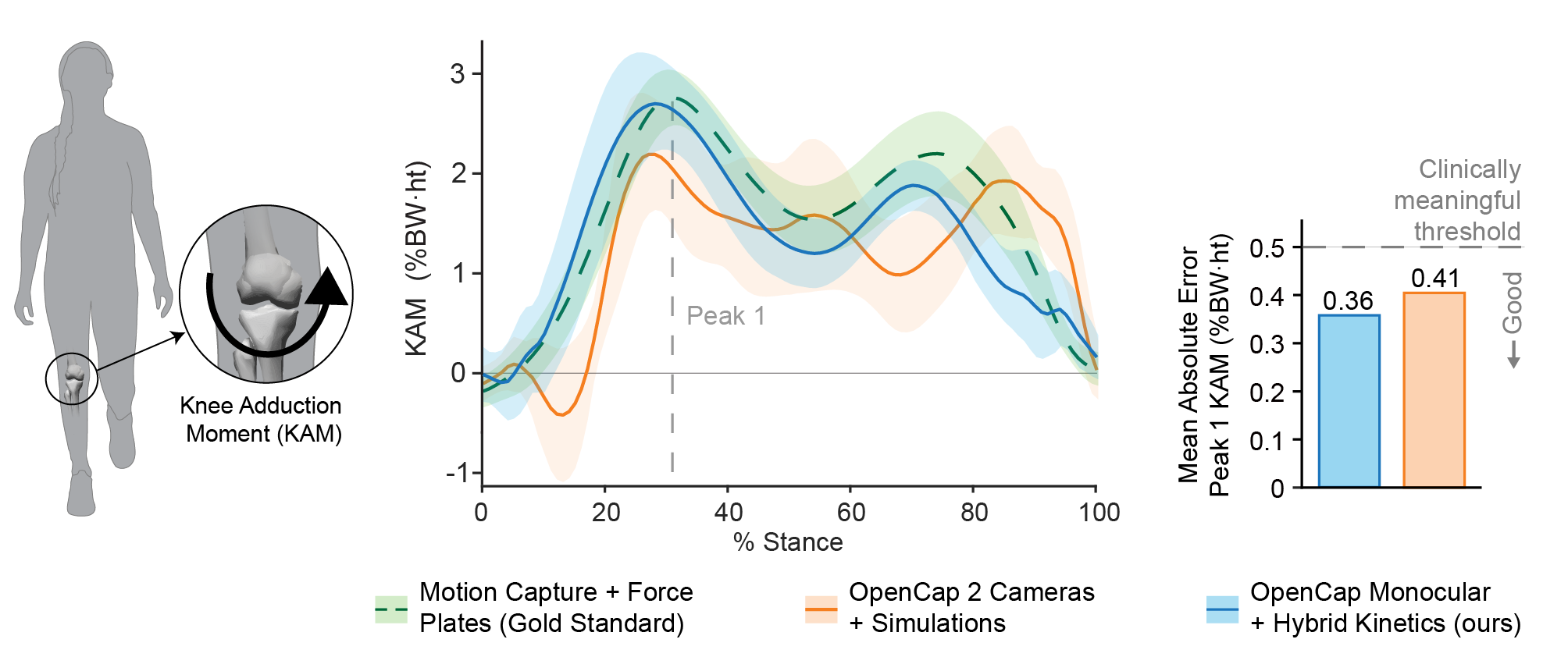}
\caption{\textbf{Clinical Use Case 2: Knee Loading during Walking.} 
The knee adduction moment (KAM) predicts the progression of medial compartment knee osteoarthritis but is traditionally difficult to measure clinically.  
(\textbf{A.}) The mean (line) and standard deviation (shading) of the KAM, normalized to bodyweight (BW) and height (ht), over the stance phase. 
(\textbf{B.}) Mean absolute error (MAE) of the first peak KAM, which is a target for biomechanical interventions. OpenCap Monocular's errors are below a clinically meaningful threshold of 0.5\% BW·ht \cite{Amin, Mundermann_1, Mundermann_2, Miyazaki}.}
\label{fig:KAM}
\end{fullwidthfigure}

\section{Discussion}

In this study, we developed OpenCap Monocular, a pipeline for analyzing 3D human musculoskeletal kinematics and kinetics from a single smartphone video. We validated it against laboratory-based motion capture and force plates, and we demonstrated sufficient accuracy to support downstream clinical tasks related to osteoarthritis and frailty. Our key contribution is the integration of a 3D human pose estimation model, physics-inspired pose optimization, and a musculoskeletal simulation framework. Together, these components produce 3D kinematics and kinetics that are physically realistic and biomechanically plausible. The pipeline bridges the gap between the SMPL-based pose estimates commonly produced by computer vision models and the skeletal kinematics and musculoskeletal dynamics that are integral to biomechanics research. By deploying the workflow in the cloud, OpenCap Monocular makes 3D biomechanical analysis freely available to researchers across movement-related fields in minutes, without requiring software development expertise or high-performance computing resources. We anticipate that OpenCap Monocular's accessibility will enable novel studies of human movement in real-world settings that were previously infeasible.

A central finding of this work is the importance of the pose-refinement optimization step for both kinematic and kinetic accuracy. Directly applying inverse kinematics to the monocular computer vision pose estimates (CV+IK) resulted in large kinematic errors, particularly in global translation, due to drift and inaccurate foot-floor interaction. Our pose refinement optimization, which enforces physical constraints such as foot–floor contact, substantially improved kinematic accuracy, reducing rotational and translational error by 48\% and 69\%, respectively, compared to CV+IK. These improvements also enhanced the accuracy of kinetics. Our physics-based simulation computes ground reaction forces based on the motion of foot-mounted contact spheres relative to a ground plane \cite{Serrancoli, Simbody}, a process that is highly sensitive to unrealistic foot-floor penetration and foot sliding. The foot-floor kinematics of the CV+IK method were not sufficiently accurate to track in simulations (Fig~\ref{fig:drift_sts}). In contrast, the refined OpenCap Monocular kinematics yielded dynamically consistent simulations of sit-to-stand activity (Fig~\ref{fig:sts}). For walking, we used the GaitDynamics machine learning model to estimate ground reaction forces from whole-body kinematics, and OpenCap Monocular produced the most accurate predictions in the vertical direction—outperforming CV+IK and the two-camera OpenCap system—likely due to improved center-of-mass kinematics resulting from the pose refinement optimization step.

In addition to our pose refinement algorithm, OpenCap Monocular's improved kinematic and kinetic accuracy relative to direct monocular computer vision outputs also stems from simplifying assumptions enabled by the OpenCap web and mobile applications. Whereas WHAM estimates global motion using a potentially moving camera with unknown intrinsic parameters and unknown participant scale, OpenCap Monocular assumes a static smartphone, known camera intrinsic parameters, and known participant height. Within the existing OpenCap workflow, these assumptions introduce minimal additional burden, as we maintain a database of camera intrinsics for all iOS devices released since 2018, and participant height is collected through the web application as part of the standard workflow. Prioritizing accuracy over flexibility was intentional to support the rapid, large-scale collection of reliable biomechanics data. However, these assumptions currently limit applications such as analyzing movements from large online video databases. Future work can leverage our open-source codebase to relax these constraints and broaden applicability beyond the OpenCap acquisition pipeline.

OpenCap Monocular produces biomechanically realistic joint kinematics and musculoskeletal dynamics, advancing the utility of monocular pose estimation for biomechanical research and clinical practice. Our accuracy evaluation using these quantities from a state-of-the-art musculoskeletal model is more informative for biomedical applications than mean per-joint position errors, which are typically used to benchmark computer vision models. Importantly, going beyond motion and estimating measures of musculoskeletal dynamics—such as muscle and joint forces—is essential for studying human performance and movement-related conditions. These dynamic quantities more directly reflect neural control and the mechanical stimuli experienced by tissues, making them more relevant to injury \cite{hewett2005, Yang, Kadono, Pijnappels} and neuro-musculoskeletal pathology \cite{frailty_seko2024, Miyazaki, Neckel} than kinematics alone. We showed that OpenCap Monocular can estimate a key measure of knee loading with an accuracy sufficient to identify individuals at risk for rapid progression of medial compartment knee osteoarthritis \cite{Miyazaki, Mundermann_1, Mundermann_2, Amin, Boswell}. We also demonstrated the ability to compute the knee extension moment and changes in lower-extremity joint moments during chair rise with sufficient accuracy to distinguish individuals with and without early signs of frailty \cite{frailty_seko2024, van_der_kruk, Moreland, heijden}. 

Cloud deployment makes advanced algorithms—previously confined to biomechanics and computer vision experts—accessible to a broad community of researchers studying human movement. Several recent developments have made this possible. Computer vision models now estimate human pose with increasing accuracy, with methods such as WHAM providing fast predictions on long videos. In parallel, muscle-driven physics simulations have become fast enough for routine use \cite{falisse_rapid}; for example, a single sit-to-stand repetition can now be simulated in minutes. Large, high-quality ground-reaction-force datasets \cite{addbio} have also enabled machine learning models that estimate kinetics with accuracy comparable to or exceeding physics-based methods \cite{tan2025gaitdynamics}. Deploying these algorithms in an easy-to-use workflow democratizes access to these cutting-edge advancements in biomechanics and computer vision. For instance, using the two-camera OpenCap system, we recently partnered with neurology clinicians to collect data from individuals with rare neuromuscular diseases across fourteen states at large-scale data collection events \cite{Ruth}. Others have used similar approaches to link joint loading to cartilage outcomes outside the MRI suite \cite{Miller_MRI}. With its more straightforward setup, OpenCap Monocular enables motion assessment in even more ecologically valid environments, such as the clinic and the home.

The single-phone approach addresses time, expertise, and equipment barriers that have limited the adoption of motion capture into clinical practice and population-scale studies. While applications requiring extremely high-precision kinematics, such as pre-surgical planning for cerebral palsy, will likely continue to justify laboratory-based motion capture \cite{cp}, OpenCap Monocular can augment or replace existing low-fidelity functional outcomes that are common in clinical practice and research. Time constraints and patients' inability to independently complete assessments are among the most frequently cited reasons why physical therapists do not adopt digital health technology \cite{AcutePT_Time, PT_OutcomeMeasures}. Even our relatively simple multi-camera OpenCap system requires setting up multiple devices, calibration, and completing a static pose for model scaling before collecting movement data \cite{opencap}. With OpenCap Monocular, users can begin collecting data in less than 1 minute from the moment they log in to the application, without any calibration or scaling steps. Data collection could be easily automated for independent patient completion in the clinic. This convenience enables clinicians to quantify informative biomechanical outcomes, such as joint angles or muscle forces, during a functional activity with little additional burden than existing low-fidelity outcomes, like time to walk 10 meters.
Furthermore, single smartphone movement assessments are feasible in the home \cite{melissa}, but prior work has measured lower-fidelity outcomes, like task completion time and 2D-projected kinematics. OpenCap Monocular can quantify 3D kinematics and kinetics in the home, enabling large-scale decentralized studies of movement health, data-driven telerehabilitation, and regular monitoring of physical function in high-risk populations. The research implications of this accessibility are a shift from sparse, laboratory-based snapshots of function to regular, real-world measurements in large cohorts.

It is important to acknowledge several limitations. First, the foot-contact probabilities from WHAM influence both its initial pose estimate and our refinement step; they perform poorly during activities involving prolonged flight phases. As a result, OpenCap Monocular does not currently perform well for jumping tasks, although alternative contact-probability models could mitigate this limitation \cite{rempe2021humor}. Nevertheless, the method performs well for activities of daily living frequently studied in mobility research. Second, our validation cohort consisted of young, healthy adults; additional studies in diverse populations, including those with pathological gait or movement disorders, are needed. Third, we evaluated a single-camera configuration (45\textdegree{} anterolateral view), which we found qualitatively superior to frontal or sagittal placements, though different activities may benefit from other viewpoints. Future work should examine performance across viewpoints and sensitivity to slight variations in camera placement. As with all monocular video-based biomechanics tools, a consistent camera setup remains important due to inherent challenges in estimating out-of-plane motion. These limitations highlight the need for evaluation across more diverse populations, activities, and environments to assess generalizability.

Finally, OpenCap Monocular was designed with a modular architecture, enabling continuous improvement as computer vision and biomechanics technologies evolve. The initial 3D pose estimation module (currently WHAM) can be readily upgraded as new monocular pose estimation or foundation models become available. Because the downstream optimization, inverse kinematics, and musculoskeletal simulation components are independent of the pose estimation algorithm, OpenCap Monocular provides a flexible and extensible framework that will continue to evolve with advances in computer vision while maintaining biomechanically grounded outputs suitable for clinical and research applications.

\section{Conclusion}

We developed and validated OpenCap Monocular, a method for quantifying 3D human motion and musculoskeletal dynamics from a single smartphone video. The pipeline combines monocular pose estimation with physics-inspired optimization to estimate accurate kinematics from video, and uses musculoskeletal simulation and machine-learning models to infer dynamics from those kinematics. OpenCap Monocular demonstrated improved accuracy compared to direct computer-vision outputs while providing a more accessible alternative to both laboratory-based motion capture and multi-camera video systems. This approach has the potential to substantially broaden the use of quantitative movement analysis in real-world settings, including telemedicine and decentralized clinical trials. By making the algorithm freely available and hosting computation in the cloud, OpenCap Monocular advances the accessibility of biomechanically grounded 3D human movement analysis.

\section{CRediT Author Contributions}
Selim Gilon: Writing--original draft, Validation, Software, Methodology, Visualization, Formal analysis.
Emily Y. Miller: Writing--review \& editing, Software.
Scott D. Uhlrich: Writing--review \& editing, Conceptualization, Software, Methodology, Supervision.

\section{Acknowledgments}
 This study was funded by grants from the Myotonic Dystrophy Foundation, the Wu Tsai Human Performance Alliance Agility Project Program, and the NIH Restore Center Pilot Project Program.

\section{Declaration of competing interest}
SDU is a co-founder of Model Health, Inc., which provides markerless motion-capture technology for commercial, non-academic use. All software presented in this work is open source, integrated into the OpenCap codebase, and incorporated into the cloud-deployed OpenCap platform, which is freely available for academic research. Model Health, Inc. had no role in the study design; data collection, analysis, or interpretation; the decision to publish; or the preparation of the manuscript.

\section{Code and Data Availability}
The OpenCap Monocular source code is openly available under a permissible Apache 2.0 License at https://github.com/utahmobl/opencap-monocular. The OpenCap web and iOS applications used for data collection are accessible at https://app.opencap.ai.  
All validation experiments used the previously published, publicly available OpenCap dataset described in Uhlrich, Falisse, Kidzinski, et al., 2023 (https://simtk.org/opencap). This dataset includes synchronized multi-camera videos, marker-based motion capture, and force plate recordings. The OpenCap Monocular outputs are available at https://simtk.org/opencap-monoc.


\nolinenumbers

\newpage

\bibliography{refs}

\end{document}



\section*{Supplementary Appendix}


Rotational accuracy for each degree of freedom is reported in Fig~\ref{S1_Fig}. Optimization parameters for each activity are listed in Table~\ref{S1_table}. The pose refinement optimization procedure is described in Section~2.2.2. Example screenshots of the OpenCap mobile and web applications are shown in Fig~\ref{S2_Fig}.

\begin{fullwidthfigure}
\includegraphics[width=\linewidth]{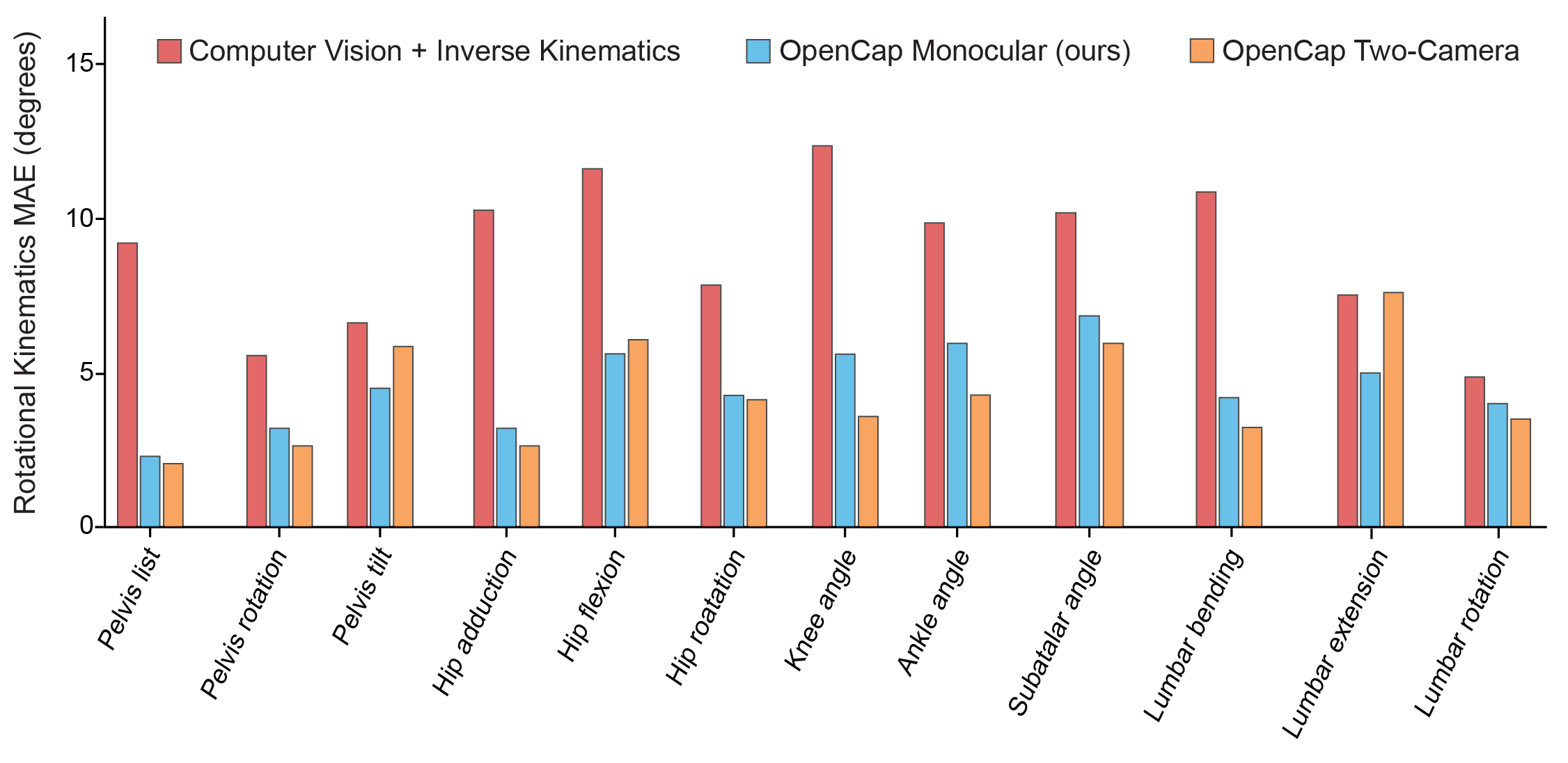}
\caption{\textbf{Rotational kinematic accuracy by degree of freedom.} Mean absolute errors (MAE) for each degree of freedom, compared to marker-based motion capture. Errors are averaged across all participants. Pose refinement in OpenCap Monocular substantially reduces errors relative to the CV+IK baseline, achieving accuracy comparable to the two-camera system.}
\label{S1_Fig}
\end{fullwidthfigure}

\begin{table}[H]
\centering
\caption{\textbf{Optimization parameters for each activity.}}
\begin{tabular}{lcccc}
\toprule
\textbf{Parameter} & \textbf{Walking} & \textbf{Squats} & \textbf{STS} & \textbf{Other} \\
\midrule
Filter frequency (Hz) & 6 & 4 & 4 & 8 \\
\midrule
Reprojection weight & 50 & 50 & 250 & 50 \\
Contact velocity weight & 1 & 1 & 1 & 1 \\
Contact position weight & 100 & 100 & 100 & 100 \\
Smoothness weight & 10 & 10 & 10 & 10 \\
Flat floor weight & 100 & 10 & 50 & -- \\
\bottomrule
\end{tabular}
\label{S1_table}
\end{table}

\begin{fullwidthfigure}
\includegraphics[width=\linewidth]{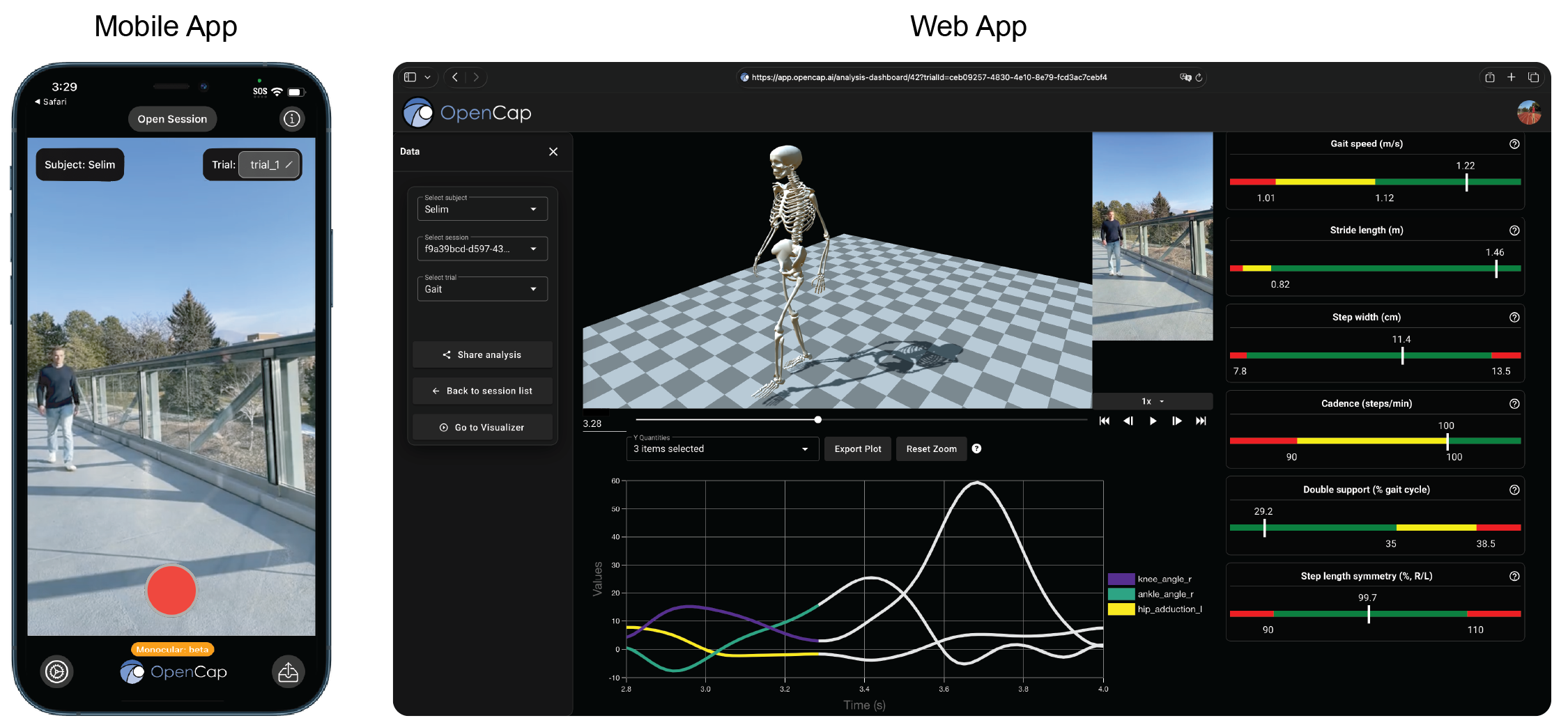}
\caption{\textbf{Screenshots of the OpenCap mobile/web app with a monocular recording.}}
\label{S2_Fig}
\end{fullwidthfigure}